\documentclass[letterpaper, 10 pt, conference]{ieeeconf}  % Comment this line out if you need a4paper

\IEEEoverridecommandlockouts                              % This command is only needed if 
\usepackage{textcomp, gensymb}
\usepackage{graphicx} 
\usepackage{amssymb}  % assumes amsmath package installed
\usepackage{amsmath} 
\usepackage{microtype}
\usepackage{graphicx}
\usepackage{amsmath}
\usepackage{amssymb}
\usepackage{booktabs}
\usepackage{algorithm}
\usepackage{algorithmic}
\usepackage{booktabs}
\usepackage{colortbl}
\usepackage[x11names]{xcolor}
\definecolor{Gray}{gray}{0.85}
\definecolor{LightCyan}{rgb}{0.88,1,1}
\usepackage{graphicx}
\usepackage{comment}
\usepackage{subfigure}

\title{ \LARGE \bf Proprioceptive State Estimation for Quadruped Robots using Invariant Kalman Filtering and Scale-Variant Robust Cost Functions }

\author{ Hilton Marques Souza Santana$^{1,2}$, João Carlos Virgolino Soares$^{2}$, Ylenia Nisticò$^{2,3}$\\ Marco Antonio Meggiolaro$^{1}$ and Claudio Semini$^{2}$% <-this % stops a space
\thanks{$^{1}$ Department of Mechanical Engineering at the Pontifical Catholic University of Rio de Janeiro,
        {\tt\small hiltonmarquess@gmail.com} /
        {\tt\small meggi@puc-rio.br}}%
\thanks{$^{2}$ Dynamic Legged Systems lab, Istituto Italiano di Tecnologia, {\tt\small joao.virgolino@iit.it}, /
{\tt\small ylenia.nistico@iit.it}, / 
{\tt\small claudio.semini@iit.it}}
\thanks{$^{3}$ Dipartimento di Informatica, Bioingegneria, Robotica e Ingegneria dei Sistemi (DIBRIS), University of Genoa, Genoa, Italy.}
}

\begin{document}

\maketitle

\begin{abstract}

Accurate state estimation is crucial for legged robot locomotion, as it provides the necessary information to allow control and navigation. However, it is also challenging, especially in scenarios with uneven and slippery terrain. This paper presents a new Invariant Extended Kalman filter for legged robot state estimation using only proprioceptive sensors. We formulate the methodology by combining recent advances in state estimation theory with the use of robust cost functions in the measurement update. We tested our methodology on quadruped robots through experiments and public datasets, showing that we can obtain a pose drift up to 40\% lower in trajectories covering a distance of over 450m, in comparison with a state-of-the-art Invariant Extended Kalman filter.

\end{abstract}

%%%%%%%%%%%%%%%%%%%%%%%%%%%%%%%%%%%%%%%%%%%%%%%%%%%%%%%%%%%%%%%%%%%%%%%%%%%%%%%%
\section{INTRODUCTION}

Quadruped robots are becoming increasingly capable, with recent advancements enabling them to perform complex tasks in unstructured environments. State estimation is a crucial step for these robots to achieve autonomy. Quadruped robots require accurate pose and velocity estimation to feed the controllers, ensure stability, and perform navigation. However, accurate state estimation is challenging due to sensor noise and external disturbances.

One of the most important aspects of state estimation is the choice of sensor configuration, which drastically impacts the performance, and applicability of the robot. State estimation frameworks can use proprioceptive (e.g., joint encoders, IMU, or torque sensors), exteroceptive sensors (e.g., cameras and LiDARs) or both. The ones that use a combination of both are usually more accurate \cite{vilens}, and have other advantages in terms of task capabilities, for instance, the ability to create maps of the environment \cite{pronto}. Recent works include VILENS \cite{vilens}, Cerberus \cite{cerberus}, and STEP \cite{step}. These works introduced a multi-sensor state estimation framework that integrates proprioceptive and exteroceptive measurements to achieve precise state estimation.

However, these sensors, particularly cameras, are prone to fail under lighting changes and scenes with low-texture. Other sensors, such as LiDARs, can fail when the robot is navigating in featureless environments such as long corridors. Furthermore, exteroceptive sensors cannot reach high frequencies and are thus subject to blur during fast motions. Thus, it is important to have a robust system that can
provide state estimates only with proprioceptive data.
\looseness=-1

\begin{figure}[t]
    \centering

  \subfigure[  \hspace{0.05cm}    Before slippage on the floor]{% 
    \includegraphics[width=.469\linewidth]{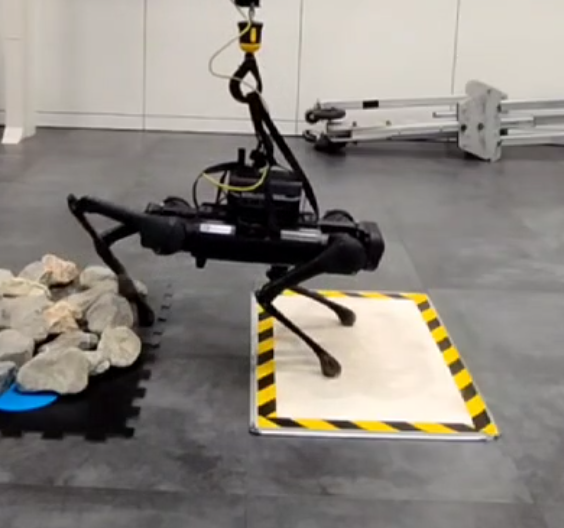}
    \label{fig:3} 
  } 
  %\quad 
  \subfigure[ \hspace{0.05cm}   Slippage on the floor]{% 
    \includegraphics[width=.473\linewidth]{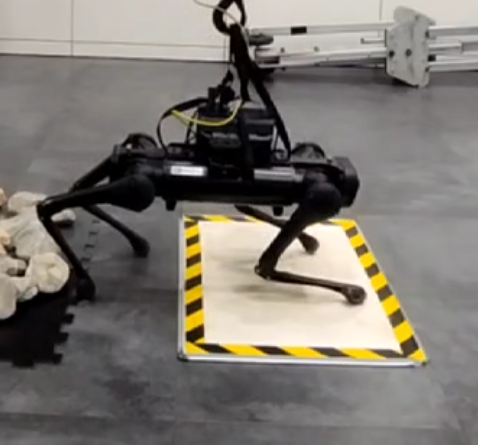}
    \label{fig:4} 
  } 
  \caption{AlienGo robot walking on slippery terrain.}

  \label{fig:firstpage}
  \vspace{-2.5mm}
\end{figure}

There are several state estimation frameworks in the literature using only proprioceptive sensors. These works use either filtering \cite{bloesch2012,fink} or smoothing \cite{invariantsmoother} methods to fuse the sensors to obtain an optimal solution. Smoothing methods are in general more accurate as they use all past estimations to optimize the current state. However, this can make it less suitable for real-time applications. Filtering methods, on the other hand, are typically more efficient, as data is processed sequentially. 
\looseness=-1

One of the first filtering-based state estimators for legged robots was proposed by Bloesch et al. \cite{bloesch2012}, based on the Extended Kalman filter (EKF), using quaternions to represent rotations, being also called QEKF. Fink and Semini~\cite{fink} proposed a system with an attitude observer decoupled from the translation part, using an eXogenous Kalman Filter (XKF), also using quaternions to represent rotations. 

One of the problems with QEKF is the dependency of the error propagation model on the state estimates \cite{invekf}. If the state estimate is not accurate, the linearization may not be a good approximation, leading to significant errors in the prediction and update steps. As the filter continuously updates the state estimate based on the error propagation model, any inaccuracies in the state estimate can accumulate over time. This can cause the error covariance to become increasingly inaccurate, degrading the overall performance of the filter.
A second limitation is that the rotation is handled independently of the position and velocity. Thus, the uncertainty propagation does not capture the coupling between
rotation and translation. 

The Invariant Extended Kalman Filter (IEKF), first proposed by Barrau and Bonnabel in \cite{invekf}, and later generalized in \cite{geometry_barrau}, solves these problems by modeling the state as a Lie group, and performing on-manifold filtering. Several variations of the IEKF have been proposed afterward \cite{hartley2020ijrr,lin2021legged,teng2021icra,lin2023proprioceptive}, with Hartley et al. being the first ones to implement the IEKF on legged robots \cite{hartley2020ijrr}. 
To handle complex scenarios, particularly when slippage occurs, Teng et al. \cite{teng2021icra} proposed an IEKF that fuses leg kinematics with inertial-velocity measurements from a tracking camera.

Despite being accurate, some of the proprioceptive estimators were not designed to handle complex scenarios such as slippery or rough terrain. Some suffer from high drift caused by slippage \cite{hartley2020ijrr}, others do not explicitly handle slippage in their formulation and experiments 
\cite{fink}, \cite{camurri2017ral}. Also, the high drift in pose usually makes proprioceptive state estimators unfeasible for use in autonomous navigation, especially considering the drift in the vertical direction, as shown by~\cite{lin2021legged}.

\subsection{Motivation}

Contact estimation is a crucial information for legged robots, especially without any exteroception. Most state estimation approaches for legged robots are tailored for rigid contact scenarios and overlook the physical characteristics of the terrain. 
This means that one of the main challenges of proprioceptive state estimation for legged robots is the drift caused by wrong contact estimation \cite{fahmi2021state}.
This can be caused by several issues that affect the legs, mainly slippage, deformable terrain, and deformable legs. 

Nistic\`o et al. \cite{ylenia2022} proposed a slip detection approach based on leg kinematics, that makes use of velocity and position measurements at the ground contacts. 
The authors detect a slippage event when the differences between the desired foot position and velocity, and the observed ones, overcome a predetermined threshold.
Camurri et al. \cite{camurri2017ral} proposed a probabilistic contact estimation and impact detection framework for quadruped robots. Their approach, however, is dependent on the gait and it does not explicitly handle the slippage problem. Yoon et al. \cite{invariantsmoother} proposed the only smoothing approach using the invariant property to date. They also dealt with the contact estimation problem, but they did not evaluate the drift in the vertical direction. We, on the other hand, evaluate the full pose drift.

Lin et al. \cite{lin2021legged} proposed a learning approach for contact estimation, showing that the traditional implementations of the IEKF, using thresholds on the ground reaction forces to determine contact, are prone to high drifts in the vertical direction. They achieved results with similar accuracy to visual odometry frameworks. However, the uncertainty of the contact estimation was not modeled in their network, leading to a potential error under high foot slippage. The same authors later used this approach as an optional contact estimation module in \cite{lin2023proprioceptive}.

Robust statistical techniques can be used to properly address outliers in contact estimation. For instance, in \cite{bloesch2013iros} Bloesch et al. use a technique called Mahalanobis distance rejection to remove the outliers. This technique, however, can cause an increase in estimation variance for discarding information contained in the residual \cite{karlgaard}.

Gao et al. \cite{Gao2022} proposed an IEKF combined with the adaptative filter proposed by Moghaddamjoo and Kirlin \cite{Moghaddamjoo}, also adding a robust cost function in the propagation model. It was observed that the average error in estimating the speed of the foot in contact decreased by 50\%.  However, they do not expĺicit which functions they use and do not analyze the impact of their values on the overall results. Furthermore, they do not analyze the impact of their methodology on the trajectory error. We, on the other hand, not only evaluate the drift in pose, but also perform an analysis of the influence of cost functions in the result.

\subsection{Contributions}

We propose a proprioceptive state estimator for quadruped robots that combines the properties of the IEKF and scale-variant robust cost functions to decrease the pose drift error caused by external disturbances to the robot.

The main contributions of the paper are listed as follows.
\begin{itemize}

    \item  We present a robust IEKF for quadruped robots, giving a new derivation of the right-invariant IEKF based on the novel work from Barrau and Bonnabel \cite{geometry_barrau}, and combining it with scale-variant robust cost functions. To the best of our knowledge, this is the first paper to use a robust filter in the measurement model of the IEKF in quadruped robot state estimation to deal with the drift caused by perturbations in the leg.
    
    \item We show the superior performance of our robust IEKF in comparison with the state-of-the-art IEKF proposed by Hartley et al. \cite{hartley2020ijrr} in terms of absolute pose accuracy, achieving up to 40\% less pose drift in trajectories covering a distance of 450m.
        
        \item A detailed analysis of the influence of the cost functions in the state estimation accuracy is also presented. We show the results obtained using the public dataset \textit{Street} from \cite{cerberus}, collected on the robot A1, and during an experiment conducted in the DLS lab (Genoa, Italy) on an AlienGo robot walking on rocks and slippery terrain.

\end{itemize}

The rest of the paper is organized as follows: Section \ref{sec:background} gives the mathematical background for the development of the proposed method, including Lie Groups and robust cost functions. In Section \ref{sec:methodology} we explain the proposed IEKF with robust cost functions. In Section \ref{sec:results} we show the results from tests with a dataset and from an experiment with a quadruped robot. Section \ref{sec:conclusion} concludes the paper.

\section{BACKGROUND}
\label{sec:background}

This section presents the mathematical background required for understanding the methodology, including Lie Groups and scale-variant robust cost functions.

\subsection{Lie Groups}
A Lie group $\mathcal{G}$ is a homogeneous differentiable manifold whose elements satisfy an algebraic structure of a group \cite{Sol2018AML}. Its algebraic structure can be summarized by the following two operations:
\begin{equation}
\begin{aligned}
	\text{Composition}: \mathcal{G} \times \mathcal{G} \to \mathcal{G}, \mathcal{X} \circ \mathcal{Y} \to \mathcal{W} 
\end{aligned},
\end{equation}
\begin{equation}
\begin{aligned}
\text{Inverse}: \mathcal{G} \to \mathcal{G}, 
\mathcal{X} \to \mathcal{X}^{-1}| \mathcal{X}^{-1} \circ \mathcal{X} = \mathcal{X} \circ \mathcal{X}^{-1} = \mathcal{E}
\end{aligned},
\end{equation}
where $\mathcal{E}$ is a distinguished element of $\mathcal{G}$ called the identity. In the rest of the work, the composition operation will be denoted only by the juxtaposition of the elements. 
The most important feature of a Lie group is its associated Lie algebra $\mathfrak{g}$, which is a vector space, and thus simpler than $\mathcal{G}$. Geometrically, $\mathfrak{g}$ is represented by the tangent space at the identity $\mathcal{T}_{\mathcal{E}}$. As described in \cite{stillwell2008}, every Lie algebra can be faithfully represented by a space of matrices. Let $\{E_{i}\}$ be a basis of $\mathfrak{g}$ and $m=\dim(\mathfrak{g})$. We now define two important isomorphisms between $\mathfrak{g}$ and $\mathbb{R}^{m}$ \cite{Sol2018AML}:
\begin{equation}
\begin{aligned}
	\text{Hat}: \mathbb{R}^{m} \to \mathfrak{g}, 
	\mathbf{v} \to \mathbf{v}^{\wedge} =  \sum_{i=1}^{m}v_{i}E_{i},
\end{aligned}
\end{equation}
\begin{equation}
\begin{aligned}
\text{Vee}: \mathfrak{g} \to \mathbb{R}^{m}, 
\mathbf{v}^{\wedge} \to (\mathbf{v}^{\wedge})^{\vee} = \mathbf{v} = \sum_{i=1}^{m} v_{i}\mathbf{e}_{i},
\end{aligned}
\end{equation}
where $\{ \mathbf{e}_{i}\}$ is a basis for $\mathbb{R}^{m}$. In this work, we extensively use the $\mathbb{R}^{m}$ representation of $\mathfrak{g}$. Furthermore, $\mathfrak{g}$ and $\mathcal{G}$ are related by the following local homeomorphism \cite{stillwell2008}:
\begin{equation}
\label{eq:exp_map}
\begin{aligned}
	\exp: \mathbb{R}^{m} \to \mathcal{G}, 
	\mathbf{v} \to  \exp(\mathbf{v}^{\wedge}) = \sum_{i=0}^{\infty} \frac{{\mathbf{v}^{\wedge}}^{i}}{i!}
\end{aligned},
\end{equation}
with its inverse defined as $\log:\mathcal{G}\to \mathbb{R}^{m}$. Following \cite{Sol2018AML}, we also overloaded the following operations:
\begin{equation}
\label{eq:main_operators}
\begin{aligned}
	\text{left-}\oplus: \mathcal{X} \times \mathcal{T}_\mathcal{X} \to  \mathcal{G},\mathcal{X} \oplus \mathbf{u} &= \mathcal{X} \exp(\mathbf{u}), \\
	\text{right-}\oplus: \mathcal{T}_\mathcal{E} \times \mathcal{X} \to \mathcal{G}, \mathbf{v} \oplus \mathcal{X} &= \exp(\mathbf{v})\exp(\mathcal{X}). \\
\end{aligned}
\end{equation}
In general, as $\mathcal{G}$ is non-commutative, we have that $\exp(\mathbf{a}) \exp (\mathbf{b}) \neq \exp(\mathbf{a} + \mathbf{b})$. Indeed, we have \cite{geometry_barrau}:
\begin{equation}
\label{eq:bch}
\exp(\mathbf{a})\exp(\mathbf{b}) = \exp(\mathbf{a} + \mathbf{b} + O(\|\mathbf{a}\|^2, \|\mathbf{a}\| \|\mathbf{b}\|, \|\mathbf{b}\|^2 )),
\end{equation}
and $\exp(\mathbf{a})\exp(\mathbf{b}) \approx \exp(\mathbf{a} + \mathbf{b})$ only if $\mathbf{a}$ and $\mathbf{b}$ are small. Another useful approximation
for the case $\mathbf{b} \in  \mathcal{T}_{\exp(\mathbf{a})}$ is using the right-Jacobian, $J_{r}(\mathbf{a})$, which allows us to write ~\cite{Sol2018AML}:
\begin{equation}
\label{eq:right_jacobian}
\exp(\mathbf{a} + \mathbf{b}) \approx \exp(\mathbf{a}) \oplus J_r(\mathbf{a}) \mathbf{b}.
\end{equation}
An important characteristic of $\mathcal{G}$ is its homogeneity, which 
ensures that the structure of the tangent space is the same everywhere \cite{Sol2018AML}. Indeed, any tangent space $\mathcal{T}_\mathcal{X}$ 
is isomorphic to $\mathcal{T}_\mathcal{E}$ by the adjoint map:
\begin{equation}
\label{eq:adjoint_map}
\text{Ad}_\mathcal{X} :\mathcal{T}_\mathcal{X} \to  \mathcal{T}_\mathcal{E},	\mathbf{u} \to \text{Ad}_\mathcal{X}\mathbf{u} =  (\mathcal{X} \mathbf{u}^{\wedge} \mathcal{X}^{-1})^{\vee}.	
\end{equation}
The main motivation behind  (\ref{eq:adjoint_map}) is to ensure the following equality \cite{Sol2018AML}:
\begin{equation}
\label{eq:adjoint}
\mathcal{X} \oplus  \mathbf{u} = \text{Ad}_\mathcal{X}\mathbf{u} \oplus \mathcal{X},
\end{equation}
which translates a local tangent vector at the frame of $\mathcal{X}$ into a global tangent vector at the frame of $\mathcal{E}$.

Finally, it is important to define errors between two trajectories on $\mathcal{G}$. Let $\mathcal{X}(t)$ be some reference trajectory and $\bar{\mathcal{X}}(t)$ its estimate. At instant $t_i$, 
it is possible to define two errors: the right-invariant $\delta \mathcal{X}_{r,i}$ and the left-invariant $\delta \mathcal{X}_{l,i}$ givens as follows \cite{hartley2020ijrr}:
\begin{equation}
\label{eq:right_error}
\delta \mathcal{X}_{r,i} := {\mathcal{X}}_{i} \bar{\mathcal{X}}_{i}^{-1} \implies \mathcal{X}_i = \boldsymbol{\xi}_{r,i} \oplus \overline{\mathcal{X}}_i    , \\
\end{equation}
\begin{equation}
\delta \mathcal{X}_{l,i} := \bar{\mathcal{X}}_{i}^{-1} \mathcal{X}_{i} \implies \mathcal{X}_i = \overline{\mathcal{X}}_i \oplus \boldsymbol{\xi}_{l,i},
\end{equation}
where $\boldsymbol{\xi}_{r,i} := \log(\delta \mathcal{X}_{r,i})$ and $\boldsymbol{\xi}_{l,i} := \log(\delta \mathcal{X}_{l,i})$. Both errors are related by the adjoint of $\bar{\mathcal{X}}_{i}$:
\begin{equation}
\label{eq:errors_adjoint}
\begin{aligned}
\mathcal{X}_i  = \boldsymbol{\xi}_{r,i} \oplus  \bar{\mathcal{X}}_i =
\bar{\mathcal{X}}_i \oplus \boldsymbol{\xi}_{l,i}  \implies
\boldsymbol{\xi}_{r,i} = \text{Ad}_{\bar{\mathcal{X}}}  \boldsymbol{\xi}_{l,i}
.\end{aligned}
\end{equation}
Intuitively, both are measuring the same error 
but the right-invariant is defined at $\mathcal{E}$ and the left-invariant is defined at $\overline{\mathcal{X}}_i$.

\subsection{Scale-Variant Robust Cost Functions}

\begin{figure}
 	\centering
 	\includegraphics[width=0.40\linewidth]{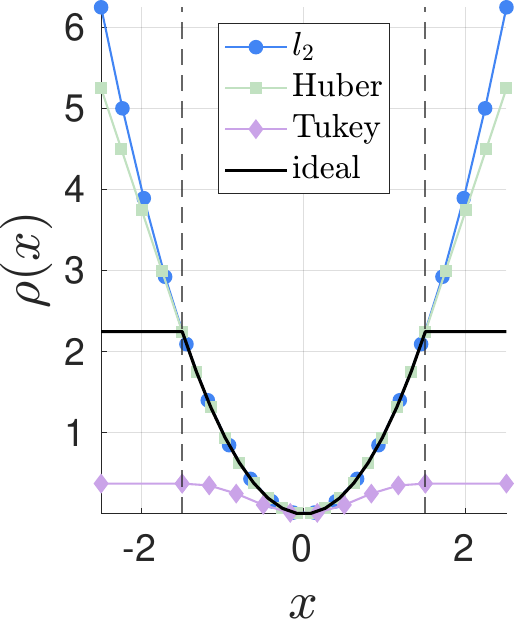}
          \caption{Comparison of the $l_2$, Huber, Tukey and ideal cost functions for $c=1.5$.}
  	\label{fig:robust_distributions}
\end{figure}

In the present work, we use scale-variant robust cost functions to deal with outliers in the measurement update of the IEKF. The main idea is to replace the $l_2$ norm used in the conventional linear regression by robust functions. These robust functions give less weight to measurements larger than a given constant $c$, a scale parameter, thus influencing less the regression. 
In this work, we tested two distinct cost functions: the Huber function (\ref{eq:huber}) and the Tukey function (\ref{eq:tukey}) \cite{zoubir2018}. The Huber cost function was chosen due to its optimal property when applied to contaminated Gaussian densities \cite{karlgaard}, while the Tukey cost function gives no weight for values above $c$, i.e., it deals better with extreme outliers. Both functions are compared with the $l_2$ function in Fig. \ref{fig:robust_distributions}, with $c = 1.5$.
\begin{equation}
\label{eq:huber}
\rho_{\text{huber}}(x) = 
\begin{cases} 
	(x - x_i)^2 & |x - x_i| \leq c \\
	c(2|x - x_i| - c) & |x - x_i| > c 
\end{cases},
\end{equation}
\begin{equation}
\label{eq:tukey}
\rho_{\text{tukey}}(x) =
\begin{cases}
\frac{c^{2}}{6}\left( 1 - \left( 1 - \frac{(x - x_{i})^{2}}{c^{2}} \right)^{3} \right), |x - x_{i}| \leq c \\
\frac{c^{2}}{6}, |x - x_{i}| > c 
\end{cases}
\end{equation}

\section{METHODOLOGY}
\label{sec:methodology}

In this Section, we present the mathematical formulation of the proposed method, consisting of the IEKF and its robust version.
\looseness=-1

\subsection{Invariant Extended Kalman Filter}

\subsubsection{Robot Model}

\begin{figure}
 	\centering
 	\includegraphics[width=0.75\linewidth]{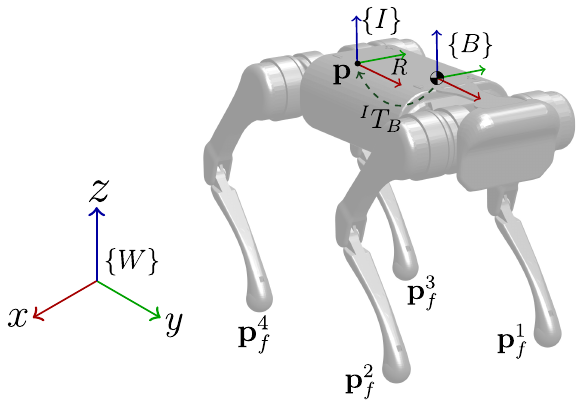}
          \caption{Main state variables and reference frames of the robot: $\mathbf{p}$ (IMU centre), $R$ (IMU orientation), $\mathbf{p}_f^{i},i=\{ 1,2,3,4\}$ (feet contact position), fixed frame ($W$), IMU frame ($I$) and base frame ($B$).}
  	\label{fig:robot_model}
\end{figure}

Figure \ref{fig:robot_model} shows the main frames of the robot considered in the method, consisting of a fixed world frame $W$, a base frame $B$, and an IMU frame $I$ and ${}^{I}T_{B}$ is a transformation from the base to the IMU frame. In our work, we use the subscript $I$ exclusively to denote vectors in the IMU frame. Vectors in the world frame do not have the subscript $W$. We assume that boolean contact information is provided.

\subsubsection{State Variables}
Following \cite{hartley2020ijrr} we define the state variables at instant $t_i$ as:
\begin{equation}
\mathcal{X}_{i} := 
\begin{bmatrix}
R_{i} & \mathbf{v}_{i} & \mathbf{p}_{i} & \mathbf{p}_{f}^{1} & \dots & \mathbf{p}_{f}^{N} \\
\mathbf{0}_{1,3} & 1 & 0 & 0 & \dots & 0 \\
\mathbf{0}_{1,3} & 0 & 1 & 0 & \dots & 0 \\
\mathbf{0}_{1,3} & 0 & 0 & 1 & \dots & 0 \\
\mathbf{0}_{1,3} & 0 & 0 & 0 & \dots & 1
\end{bmatrix} \in SE_{2+N}(3)
\end{equation}
where $R_i \in SO_3$, $\mathbf{p} \in  \mathbb{R}^{3}$, $\mathbf{v}_i \in  \mathbb{R}^{3}$ are the orientation, position, and velocity of the IMU in the world frame, 
$\mathbf{p}_{f}^{i} \in  \mathbb{R}^{3}$, $i=\{1,2,3,4\}$, is the contact position of the i-\textit{th} foot also in the world frame, and $N$, $1 \leq N \leq 4$, represents the
number of legs that are in contact at instant $t_i$.

The Lie group 
$SE_{2+N}(3) := SO_3 \ltimes \mathbb{R}^{6 + 3N}$, $SO_3$ being the special orthogonal group, was initially proposed in \cite{barrau2015ekf} and was called there the ``group of multiple direct spatial isometries".
We also make use of the following notation for representing $\mathcal{X}_i$ \cite{geometry_barrau}:
\begin{equation}
\mathcal{X}_i := (R_i, \mathbf{x}), 
\mathbf{x} := \begin{bmatrix} \mathbf{v}_i, \mathbf{p}_i, \mathbf{p}_f^{1}, \ldots, \mathbf{p}_f^{N} \end{bmatrix}^{T}
\end{equation}
Then, the composition rule can be given as:
\begin{equation}
\label{eq:composition_rule}
\mathcal{X}_i	\mathcal{X}_j = (R_i, \mathbf{x}_i) (R_j, \mathbf{x}_j) = (R_iR_j, \mathbf{x}_i + R_i * \mathbf{x}_j),
\end{equation}
where $(*)$ denotes a term-by-term multiplication, e.g., 
$R_i * \begin{bmatrix} \mathbf{x}_1,\ldots, \mathbf{x}_n \end{bmatrix}^{T} = \begin{bmatrix} R_i \mathbf{x}_1, \ldots, R_i \mathbf{x}_n \end{bmatrix}^{T}$ \cite{geometry_barrau}. 
For the $SE_{2+N}(3)$ the exponential map (\ref{eq:exp_map}) is defined as:
\begin{equation}
\begin{aligned}
	\exp(\mathbf{v}) = \exp(\begin{bmatrix} \boldsymbol{\xi}^{R} & \boldsymbol{\xi}^{\mathbf{x}}  \end{bmatrix}^{T} ) = (\underbrace{\exp_{SO_3}(\boldsymbol{\xi}^{R})}_{R}, 
	\underbrace{J_{l}(\boldsymbol{\xi}^{R}) \boldsymbol{\xi}^{\mathbf{x}}}_\mathbf{x})
\end{aligned}
,
\end{equation}
where $\mathbf{v}^{\wedge} \in  \mathfrak{se}_{2+N}(3)$, $J_l$ is the left-Jacobian of $SO_3$ \cite{Sol2018AML} and $\exp_{SO_3}$ 
the exponential map of the $SO_3$ \cite{Sol2018AML}. The adjoint map (\ref{eq:adjoint_map}) is given by \cite{hartley2020ijrr}:
\begin{equation}
\label{eq:adjoint_se3n}
\text{Ad}_{\mathcal{X}} = 
\begin{bmatrix}
R & \mathbf{0}_{3} &  \dots & \mathbf{0}_{3} \\
\mathbf{v}^{\times} R & R  & \dots & \mathbf{0}_{3} \\
\vdots & \vdots  & \ddots &  \vdots\\
\mathbf{p}_{f}^{\times,N}R & \mathbf{0}_{3} & \dots & R
\end{bmatrix} := 
\begin{bmatrix}
R & \mathbf{0}_{3,3(N+2)} \\
\mathbf{x}^{\times}* R & \boldsymbol{R}
\end{bmatrix}
.
\end{equation}
where $\boldsymbol{R}:=\text{diag}(R, \underbrace{ \dots }_{ N }, R)$ and $(\cdot)^{\times}$ is the skew-symmetric operator \cite{Sol2018AML}. Furthermore, following the rule proposed in \cite{geometry_barrau}, we only focus on the present work in the right-invariant error as our observation (\ref{eq:measurement}) is performed in the IMU frame (body frame in \cite{geometry_barrau}). 

For simplicity, in the following sections, we only consider one leg in contact, i.e., $\mathcal{X}_{i} = (R_{i}, \begin{bmatrix}
\mathbf{v}_{i} & \mathbf{p}_{i} & \mathbf{p}_{f,i}
\end{bmatrix}^{T}) \in SE_{3}(3)$. The strategy for updating the state when switching legs in contact is the same as in \cite{hartley2020ijrr}.

\subsubsection{Discrete Stochastic Process Model}
In this work, we assume that biases of the IMU are known, and the measured angular velocity $\tilde{\boldsymbol{\omega}}_{I,i}$ and linear acceleration $\tilde{\mathbf{a}}_{I,i}$ 
are corrupted by white Gaussian noise:
\begin{equation}
\begin{aligned}
	\tilde{\boldsymbol{\omega}}_{I,i} = \boldsymbol{\omega}_{I,i} + \mathbf{w}_{g,i}, \mathbf{w}_{g,i} \sim \mathcal{N}(\mathbf{0}, Q_g) \\
	\tilde{\mathbf{a}}_{I,i} = \mathbf{a}_{I,i} + \mathbf{w}_{a,i}, \mathbf{w}_{a,i}  \sim \mathcal{N}(\mathbf{0}, Q_a)
.\end{aligned}
\end{equation}
Moreover, we also assume that the contact foot position remains fixed in the world frame. To accommodate potential slippage, 
we assume that its velocity is corrupted by white Gaussian noise defined in the IMU frame \cite{bloesch2012}: 
\begin{equation}
\label{eq:bloesch_noise}
\dot{\mathbf{p}}_{f,i} = R_i \mathbf{w}_{f,i}, \mathbf{w}_{f,i} \sim \mathcal{N}(\mathbf{0}, Q_f).
\end{equation}
\noindent
Let $\overline{\mathcal{X}}_i$ be the nominal state at instant $t_i$, the discrete evolution dynamics of $\overline{\mathcal{X}}_i$,
with no approximations, is given by \cite{hartley2020ijrr}:
\begin{equation}
\label{eq:dynamics}
\begin{aligned}
    &\overline{\mathcal{X}}_{i+1|i} = \mathcal{F}(\overline{\mathcal{X}}_i, \tilde{{\mathbf{u}}}_{i+1}, \mathbf{w}_i, \Delta t) = \\
&\begin{bmatrix}
\bar{R}_{i}\exp(\tilde{\boldsymbol{\omega}}_{I,i+1} + \Delta t) \\
\bar{\mathbf{v}}_{i} + \mathbf{g}\Delta t + \bar{R}_{i} \Gamma_{1}(\tilde{\boldsymbol{\omega}}_{i+1}\Delta t)\tilde{\mathbf{a}}_{I,i+1} \\
\bar{\mathbf{p}}_{i} + \bar{\mathbf{v}}_{i}\Delta t + \mathbf{g}\frac{\Delta t^{2}}{2} + \bar{R}_{i}\Gamma_{2}(\tilde{\boldsymbol{\omega}}_{i+1} \Delta t)(\tilde{\mathbf{a}}_{I,i+1}) \\
\bar{\mathbf{p}}_{f,i} + \bar{R}_{i}\mathbf{w}_{f,i} \Delta t
\end{bmatrix},
\end{aligned}
\end{equation}
where $\Gamma_{n}(\boldsymbol{\omega},\Delta t) = \frac{\sum_{k=0}^{\infty} \boldsymbol{\omega}^{\wedge k}\Delta t ^{k + n}}{(k+n)!}$ \cite{sola2017} and
\begin{equation}
\begin{aligned}
&\tilde{\mathbf{u}}_{i+1}  := 
\begin{bmatrix} \boldsymbol{\omega}_{I,i} + \tilde{\mathbf{w}}_{g,i+1} \\ \mathbf{a}_{I,i} + \tilde{\mathbf{w}}_{a,i+1} \end{bmatrix},
\\
&\mathbf{w}_i = 
\begin{bmatrix} \mathbf{w}_{g,i} \\ \mathbf{w}_{a,i} \\ \mathbf{w}_{f,i} \end{bmatrix} 
\sim \mathcal{N}(\mathbf{0}, Q_i), Q_{i} := 
\begin{bmatrix}
Q_{g} & \mathbf{0}_{3} & \mathbf{0}_{3} \\
\mathbf{0}_{3} & Q_{a} & \mathbf{0}_{3} \\
\mathbf{0}_{3} & \mathbf{0}_{3} & Q_{f} 
\end{bmatrix}
.\end{aligned}
\end{equation}
\noindent
We introduce the following approximations \cite{sola2017}:
\begin{equation}
\label{eq:approximations}
\begin{aligned}
&\Gamma_{1}(\tilde{\boldsymbol{\omega}}_{I,i+1}\Delta t) \approx   I_3\Delta t  \\
&\Gamma_{2} (\tilde{\boldsymbol{\omega}}_{I,i+1}\Delta t) \approx  I_3\frac{\Delta t ^{2}}{2} \\
&\exp(\tilde{\boldsymbol{\omega}}_{I,i+1}\Delta t) \stackrel{\ref{eq:right_jacobian}}{\approx}\\&\exp(\boldsymbol{\omega}_{I,i+1}\Delta t)\exp(-J_{r}(\tilde{\boldsymbol{\omega}}_{I,i+1}\Delta t) \mathbf{w}_{g,i+1})
,\end{aligned}
\end{equation}
where $J_r$ is the right-Jacobian of the  $SO_3$ \cite{Sol2018AML}. Then, (\ref{eq:dynamics}) can be rewritten as:
\begin{equation}
\label{eq:dynamics_approximated}
\begin{aligned}
&\overline{\mathcal{X}}_{i+1} = \mathcal{F}(\overline{\mathcal{X}}_i, {\tilde{\mathbf{u}}}_{i+1}, \Delta t) \approx \\
&\begin{bmatrix}
\bar{R}_{i}\exp({\boldsymbol{\omega}}_{I,i+1}\Delta t) \exp(G_{i+1}^{1}\mathbf{w}_{g,i+1}) \\
\bar{\mathbf{v}}_{i} + \mathbf{g}\Delta t + \bar{R}_{i} \tilde{\mathbf{a}}_{I,i+1} \Delta t \\
\bar{\mathbf{p}}_{i} + \bar{\mathbf{v}}_{i}\Delta t +  \mathbf{g}\frac{\Delta t^{2}}{2}+ \bar{R}_{i}\tilde{\mathbf{a}}_{I,i+1}\frac{\Delta t^{2}}{2} \\
\bar{\mathbf{p}}_{f,i} + \bar{R}_{i}\mathbf{w}_{p_{f}} \Delta t
\end{bmatrix} =  \\
&\begin{bmatrix}
	\bar{R}_{i} \exp(\boldsymbol{\omega}_{I,i+1}\Delta t) \exp(G_{i+1}^{1}\mathbf{w}_{g,i+1}\Delta t) \\
F_{i}\mathbf{x}_{i} + \mathbf{d}_{i} + R_i * \mathbf{s}_{i} + R_{i} *  G^{2}_{i+1}[\mathbf{w}_{a},\mathbf{w}_{f}]_{i+1}^{T} \end{bmatrix} = \\
&\mathcal{W}_{i}\Phi_{i}(\bar{\mathcal{X}}_{i}){\bar{\mathcal{Y}}}_{i}(\mathbf{w}_{i}),
\end{aligned}
\end{equation}
where
\begin{equation}
\label{eq:auxliary_variables}
\begin{aligned}
&\mathcal{W}_i = (I_3, \mathbf{d}_i), \\
&\bar{\mathcal{Y}}_i = (\exp({\boldsymbol{\omega}}_{I,i+1}\Delta t) \exp(G_{i+1}^{1} \mathbf{w}_{g}), \mathbf{s}_{i} + G_{i+1}^{2} 
[\mathbf{w}_{a},\mathbf{w}_{f}]_{i+1}^{T}), \\
&\Phi_i(\overline{\mathcal{X}}_{i})= (R_i, F_i \mathbf{x}_i) \\
&	F_{i} = \begin{bmatrix}
I_3 & \mathbf{0}_{3} & \mathbf{0}_{3} \\
\Delta t I_3 & I_3 & \mathbf{0}_{3} \\
\mathbf{0}_{3} & \mathbf{0}_{3} & I_{3} 
\end{bmatrix},
\mathbf{d}_{i} = \begin{bmatrix}
\Delta t \mathbf{g} \\
\frac{\Delta t^{2}}{2} \mathbf{g} \\ \mathbf{0}_{3,1}
\end{bmatrix},
\mathbf{s}_{i} = 
\begin{bmatrix}
 {\mathbf{a}}_{I,i+1}\Delta t \\
 {\mathbf{a}}_{I,i+1} \frac{\Delta t^{2}}{2}  \\
\mathbf{0}_{3,1}
\end{bmatrix}, \\
&G_{i+1}^{1} = -J_{r}(\tilde{\boldsymbol{\omega}}_{I,i+1}\Delta t), G_{i+1}^{2} =  
\begin{bmatrix}
I_3 & \mathbf{0}_3 \\
I_3\frac{\Delta t}{2} & \mathbf{0}_3  \\
\mathbf{0}_3 & I_3 
\end{bmatrix}\Delta t.
\end{aligned}
\end{equation}
Note that $\mathcal{W}_{i},\mathcal{X}_{i},\bar{\mathcal{Y}}_{i} \in SE_{3}(3)$ and (\ref{eq:dynamics_approximated}) can naturally be extended 
for when $n$ legs are in contact.
Equation \ref{eq:dynamics_approximated} is an example of \textit{natural frame dynamics} defined in \cite{geometry_barrau} which satisfies
the group affine property (Theorem 2 in \cite{geometry_barrau}). It can be shown that this property is equivalent to $\Phi$ being 
an automorphism of $SE_{3}(3)$. Every automorphism $\Phi:\mathcal{G}\to \mathcal{G}$ satisfies \cite{barrau2019linear}:
\begin{equation}
\label{eq:automorphism}
\Phi(\mathcal{X}_1\mathcal{X}_2) = \Phi(\mathcal{X}_1) \Phi(\mathcal{X}_2), \mathcal{X}_1, \mathcal{X}_2 \in G.
\end{equation}
Moreover, for every automorphism of $\mathcal{G}$ there is a linear map $M$ such that (Theorem 15 in \cite{barrau2019linear}):
\begin{equation}
\label{eq:log_linear}
\Phi(\exp(\boldsymbol{\xi})) = \exp(M \boldsymbol{\xi}), \boldsymbol{\xi}^{\wedge} \in \mathfrak{g},M \in \mathbb{R}^{m \times m}.
\end{equation}
For $\Phi$ defined as (\ref{eq:dynamics_approximated}), $M$ is given by:
\begin{equation}
M =\begin{bmatrix}
I_{3} & \mathbf{0}_{9} \\
\mathbf{0}_{9} & F
\end{bmatrix}
\end{equation}
\subsubsection{Derivation of Right-Invariant Error Propagation Without Noise}
We first consider the situation where $\mathbf{w}_i = \mathbf{0}, {\mathcal{Y}}_i := \overline{\mathcal{Y}}_i(\mathbf{0})$, and the uncertainty at $t_i$ 
is solely due to the propagation of the uncertainty in the initial position. Then, analogous to \cite{brossard22}, 
we derive the propagation of the right-invariant error:
\begin{equation}
\label{eq:discrete_log_linearity}
\begin{aligned}
	&\mathcal{X}_{i+1} = \mathcal{W}_{i} \Phi(\mathcal{X}_{i}) \mathcal{Y}_{i} \stackrel{\ref{eq:right_error}}{=}  \\
&\mathcal{W}_{i}\Phi(\boldsymbol{\xi}_{r,i} \oplus  \bar{\mathcal{X}}_{i}) \mathcal{Y}_{i} \stackrel{\ref{eq:automorphism}}{=} \\ 
&\mathcal{W}_{i} \Phi(\exp(\boldsymbol{\xi}_{r,i})) \Phi(\bar{\mathcal{X}}_{i})\mathcal{Y}_{i} \stackrel{\ref{eq:log_linear}}=  \\
&\mathcal{W}_{i} \exp(M \boldsymbol{\xi}_{r,i}) \Phi (\bar{\mathcal{X}}_{i}) \mathcal{Y}_{i} \stackrel{\ref{eq:adjoint}}{=}  \\
&\exp(\underbrace{ \text{Ad}_{\mathcal{W}_{i}} M_i \boldsymbol{\xi}_{r,i} }_{ \boldsymbol{\xi}_{r,i+1} }) \mathcal{W}_{i} \Phi(\bar{\mathcal{X}}_{i})  \mathcal{Y}_{i} \implies \\
&\boldsymbol{\xi}_{r,i+1} = \text{Ad}_{\mathcal{W}_i} M_i \boldsymbol{\xi}_{r,i} := A_i \boldsymbol{\xi}_{r,i}
\end{aligned}
\end{equation}
Equation (\ref{eq:discrete_log_linearity}) is known as the discrete log-linearity property~\cite{chauchat2023} and has two main benefits. The first one 
is that although in a non-linear space, the error can be propagated in exact form. The second one is that $A_i$ does not depend on
the current state, i.e., the propagation is autonomous. 

\subsubsection{Derivation of Right-Invariant Error Propagation With Noise}
Following \cite{geometry_barrau}, we extend the error propagation of (\ref{eq:discrete_log_linearity}) to account for the noisy system
defined in (\ref{eq:dynamics_approximated}). Let $\delta \mathcal{X}_r := (e^{R}_i, e^{\mathbf{x}}_i) \stackrel{\ref{eq:composition_rule}}{=} (R_i \bar{R}_i^{-1}, \mathbf{x}_i - R_i \bar{R}^{-1}*\bar{\mathbf{x}}_i)$. We will individually propagate 
the noisy error for the rotational and linear parts. For the rotational part we have:
\begin{equation}
\label{eq:noisy_rotational}
\begin{aligned}
&e^{R,noisy}_{i} := R_{i}^{noisy} \bar{R}_{i}^{-1} \stackrel{\ref{eq:dynamics_approximated}}{=} \\ &R_{i}\exp(G_{i+1}^{1}\mathbf{w}_{g,i+1}) \bar{R}_{i}^{-1} \stackrel{\ref{eq:adjoint}}{=} \\&R_{i} \bar{R}_{i}^{-1} \exp(\text{Adj}_{\bar{R}_{i}}G_{i+1}^{1}\mathbf{w}_{g,i+1}) \stackrel{\ref{eq:main_operators}}{=}\\&  e^{R}_{i} \oplus \text{Ad}_{\bar{R}_{i}} G_{i+1}^{1}\mathbf{w}_{g,i+1} \implies \\
&\boldsymbol{\xi}^{R,noisy} \stackrel{\ref{eq:bch}}{\approx} \boldsymbol{\xi}^{R} + \text{Adj}_{\bar{R}_{i}} G_{i+1}^{1}\mathbf{w}_{g,i+1}  
.\end{aligned}
\end{equation}
For the linear part, we have:
\begin{equation}
\label{eq:noise_1}
\begin{aligned}
&e^{\mathbf{x},noisy}_i :=  \mathbf{x}^{noisy}_i - R^{noisy} \bar{R}_{i}^{-1} * \bar{\mathbf{x}}_{i} \stackrel{\ref{eq:noisy_rotational}}{=} \\ 
&\mathbf{x}^{noisy}_i - e^{R,noisy}_{i} *\bar{\mathbf{x}}_{i} \stackrel{\ref{eq:dynamics_approximated}}{=} \\ 
&\mathbf{x}_{i} + \bar{R}_{i}*G^{2}_{i+1}[\mathbf{w}_{a},\mathbf{w}_{f}]_{i+1}^{T} - e^{R,noisy}_{i} *\bar{\mathbf{x}}_{i}.
\end{aligned}
\end{equation}
By substituting the following approximations \cite{geometry_barrau}:
\begin{equation}
\label{eq:noisy_app}
\begin{aligned}
&e^{R,noisy}_{i} = \exp(\boldsymbol{\xi}^{R,noisy}) \stackrel{\ref{eq:exp_map}}{\approx}  (I_3 + \boldsymbol{\xi}^{\times,R,noisy}), \\&
e^{\mathbf{x}}_{i} \approx \mathbf{x}_{i} - \bar{\mathbf{x}}_{i} - \boldsymbol{\xi}^{\times,R}* \bar{\mathbf{x}}_i \approx  \boldsymbol{\xi}_i^{\mathbf{x}}
\end{aligned}
\end{equation}
into (\ref{eq:noise_1}) we get the noisy propagation of the linear part:
\begin{equation}
\label{eq:linear_noisy}
\begin{aligned}
&e^{\mathbf{x},noisy} \approx \boldsymbol{\xi}^{\mathbf{x}, noisy} \approx  \\&
\mathbf{x}_{i} - \bar{\mathbf{x}}_{i} - \boldsymbol{\xi}^{\times,R} * \bar{\mathbf{x}}_i + \bar{R}_{i}*G^{2}_{i+1}[\mathbf{w}_{a},\mathbf{w}_{f}]_{i+1}^{T} - \\&
 (\text{Ad}_{\bar{R}_{i}} G_{i+1}^{1}\mathbf{w}_{g,i+1})^{\times}*\bar{\mathbf{x}}_{i} =  \\&
\boldsymbol{\xi}^{\mathbf{x}} + \bar{R}_{i}*G^{2}_{i+1}[\mathbf{w}_{a},\mathbf{w}_{f}]_{i+1}^{T} + \bar{\mathbf{x}}_{i}^{\times}*
\text{Ad}_{\bar{R}_{i}} G_{i+1}^{1}\mathbf{w}_{g,i+1},
\end{aligned}
\end{equation}
where $-\mathbf{a}^{\times}*\mathbf{x} = \mathbf{x}^{\times}*\mathbf{a}$ \cite{geometry_barrau} and  $\text{Ad}_{\bar{R}_i} = \bar{R}_i$  \cite{Sol2018AML}. Finally, the complete right-invariant error propagation is given by: 
\begin{equation}
\label{eq:error_propagation}
\boldsymbol{\xi}_{r,i+1}^{noisy} \approx A_{i}\boldsymbol{\xi}_{r,i} + B_{i} \mathbf{w}_{i+1},
\end{equation}
where
\begin{equation}
\label{eq:B_noise}
\begin{aligned}
B_{i} = 
\underbrace{ \begin{bmatrix}
\bar{R}_i   & \mathbf{0}_{3,9} \\
\bar{\mathbf{x}}_{i}^{\times} * \bar{R}_i & \bar{\boldsymbol{R}}_{i} 
\end{bmatrix}  }_{ \text{Ad}_{\bar{\mathcal{X}}_{i}} }
\underbrace{ \begin{bmatrix}
G_{i+1}^{1} & \mathbf{0}_{3,6} \\
G_{i+1}^{1}  & G_{i+1}^{2}
\end{bmatrix} }_{ G_{i+1} }
\end{aligned}.
\end{equation}
Note that only if $G_{i+1} \approx G_{i+1}^{*}=
\begin{bmatrix}
I_{3} & \mathbf{0}_{3} & \mathbf{0}_{3} \\
\mathbf{0}_{3} & I_{3} & \mathbf{0}_{3} \\
\mathbf{0}_{3} & \mathbf{0}_{3} & \mathbf{0}_{3} \\
\mathbf{0}_{3} & \mathbf{0}_{3} & I_{3}
\end{bmatrix}\Delta t$ we get $B_i = \text{Ad}_{\bar{\mathcal{X}}_i} G_{i+1}^{*}$ and ($\ref{eq:error_propagation}$) 
will be the same right-invariant error propagation obtained in \cite{hartley2020ijrr} and \cite{lin2023proprioceptive}. 
Since the error dynamics evolves as (\ref{eq:error_propagation}), its covariance propagation is given by:
\begin{equation}
\begin{aligned}
P_{i+1|i} := \mathbb{E}[\boldsymbol{\xi}_{r,i+1} \boldsymbol{\xi}_{r,i+1}^{T}] = \\ 
A_i \mathbb{E}[\boldsymbol{\xi}_{r,i}\boldsymbol{\xi}^{T}_{r,i}] A_i^{T} + B_i Q_i B_i^{T}
.\end{aligned}
\end{equation}
where we assumed $\boldsymbol{\xi}^{noisy}_r = \boldsymbol{\xi}_r$ for simplicity.

\subsubsection{Measurement Update}
The second main ingredient of invariant filtering involves representing the measurement model as a 
group action of $\mathcal{G}$ \cite{geometry_barrau}. Following \cite{bloesch2012} and \cite{hartley2020ijrr}, the measurement model is defined using forward kinematics of the contact points. 

Let $\mathbf{q} \in S_1^{12}$ denote the joint positions of the quadruped robot. We assumed
that the sampled encoder measurements at instant $t_i$ are corrupted by white Gaussian noise: 
\begin{equation}
\tilde{\mathbf{q}}_i = \mathbf{q}_i + \mathbf{w}_\mathbf{q} \sim \mathcal{N}(\mathbf{0}, Q_\mathbf{q}).
\end{equation}
Let $\mathcal{K}(\mathbf{q}):S_1^{12}\to \mathbb{R}^{3}$ denote the forward kinematics of the leg in contact. The measured foot contact
position $\tilde{\mathbf{z}}_{I,i}$ in the IMU frame at instant $t_i$ is given by \cite{bloesch2012}:
\begin{equation}
\label{eq:measurement}
\begin{aligned}
&\tilde{\mathbf{z}}_{I,i} = {}^{I}T_B\tilde{\mathcal{K}}(\tilde{\mathbf{q}}_i) \approx 
{}^{I}T_B(\tilde{\mathcal{K}}(\mathbf{q}_i) - J(\tilde{\mathbf{q}}_i)\mathbf{w}_{\mathbf{q}}) = \\&
{}^{I}T_B (\mathcal{K}(\mathbf{q}_i) + {\mathbf{w}_{\mathcal{K}} - J(\tilde{\mathbf{q}}_i)\mathbf{w}_{\mathbf{q}} })
,\end{aligned}
\end{equation}
where ${}^{I}T_B$ denotes a change of basis from the body frame to the IMU frame (Fig. \ref{fig:robot_model}), $J$ is the Jacobian of $\mathcal{K}$ and 
$\mathbf{w}_\mathcal{K} \sim \mathcal{N}(\mathbf{0}, Q_\mathcal{K})$ is a white Gaussian noise responsible for accommodating errors in the kinematics calibration.
The  final covariance of $\tilde{\mathbf{z}}_{I,i}$ is given as follows:
\begin{equation}
N_i := \mathbb{E}[\tilde{\mathbf{z}}_{I,i} \tilde{\mathbf{z}}_{I,i}^{T}] = {}^{I}T_B (Q_\mathcal{K} + J Q_\mathbf{q} J^{T}) {}^{I}T_B^{T}.
\end{equation}

We now have to describe the measurement model as a right action on the measurement space $\mathbb{R}^{3}$ \cite{geometry_barrau}.
For the forward kinematics, we have:
\begin{equation}
\begin{aligned}
	&\mathbf{z}_{I,i} = \bar{R}_{i+1|i}^{T}\left(
-\underbrace{ \begin{bmatrix}
\mathbf{0}_{3} &  {I}_{3} & -I_{3}
\end{bmatrix} }_{ H_{\mathbf{x}} } \begin{bmatrix}
\bar{\mathbf{v}}_{i+1|i} \\
\bar{\mathbf{p}}_{i+1|i} \\
\bar{\mathbf{p}}_{f,i+1|i}
\end{bmatrix}\right) = \\& \bar{R}^{T}_{i+1|i} (-H_{\mathbf{x}} \mathbf{x}) 
.\end{aligned}
\end{equation}
The relation between the innovation and $\boldsymbol{\xi}_{r,i+1|i}$ is given by (Proposition 13 in \cite{geometry_barrau}):
\begin{equation}
\delta \mathbf{z}_{i} =  \bar{R}_{i+1|i}(\tilde{\mathbf{z}}_{I,i} - \mathbf{z}_{I,i}) = \underbrace{\begin{bmatrix} \mathbf{0}_3 & -H_\mathbf{x} \end{bmatrix}}_{H_{i}} \boldsymbol{\xi}_{r,i+1|i}.
\end{equation}
Note that, unlike the usual Kalman filter, the innovation is defined in a frame opposite to that of the measurement. This feature is important because it ensures that 
the update step is state-trajectory independent (Proposition 3 in \cite{geometry_barrau}).
Finally, the estimation of the error at instant $t_i$, $\hat{\boldsymbol{\xi}}_{r,i}$, is obtained by the usual Extended Kalman Filter equations \cite{farrell2008}: 
\begin{equation}
\label{eq:aux_s}
S_i = H_i P_{i+1|i}H_i^{T} + \underbrace{\bar{R}_{i+1|i}N_i\bar{R}_{i+1|i}^{T}}_{\hat{N}_i}
,
\end{equation}
\begin{equation}
K_i:= P_{i+1|i}H_i^{T}S_i^{-1}
,
\end{equation}
\begin{equation}
\hat{\boldsymbol{\xi}}_{r,i} = K_i \delta \mathbf{z}_{i},
\end{equation}
\begin{equation}
P_{i+1} = (I - K_i H_i)P_{i+1|i}
.
\end{equation}

\begin{figure}
 	\centering
 	\includegraphics[width=1.0\linewidth]{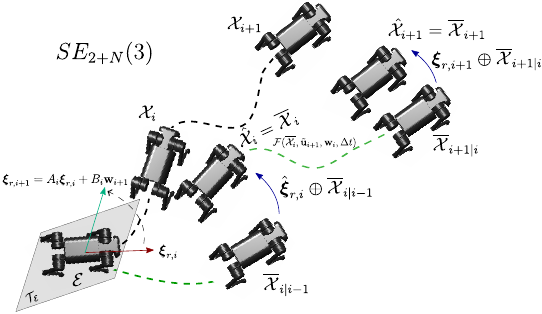}
          \caption{Summary of right-invariant error IEKF. The true state is in black, the nominal state (green), $\bar{\mathcal{X}}_i$, is propagated with (\ref{eq:dynamics_approximated}), the right-invariant error, $\boldsymbol{\xi}_{r,i}$, with (\ref{eq:error_propagation}) and the state update is given by (\ref{eq:aux_s})-(\ref{eq:state_update}).}
  	\label{fig:error_update}
\end{figure}

The current state is updated with (\ref{eq:right_error}) as follows: 
\begin{equation}
\label{eq:state_update}
\hat{\mathcal{X}}_{i} = \hat{\boldsymbol{\xi}}_{r,i+1} \oplus \overline{\mathcal{X}}_{i+1|i}.
\end{equation}
The aforementioned methodology is summarized in Fig. \ref{fig:error_update}.

\subsection{Robust Invariant Extended Kalman Filter}
\begin{figure}[t!]
    \centering
    \includegraphics[width=0.7\linewidth]{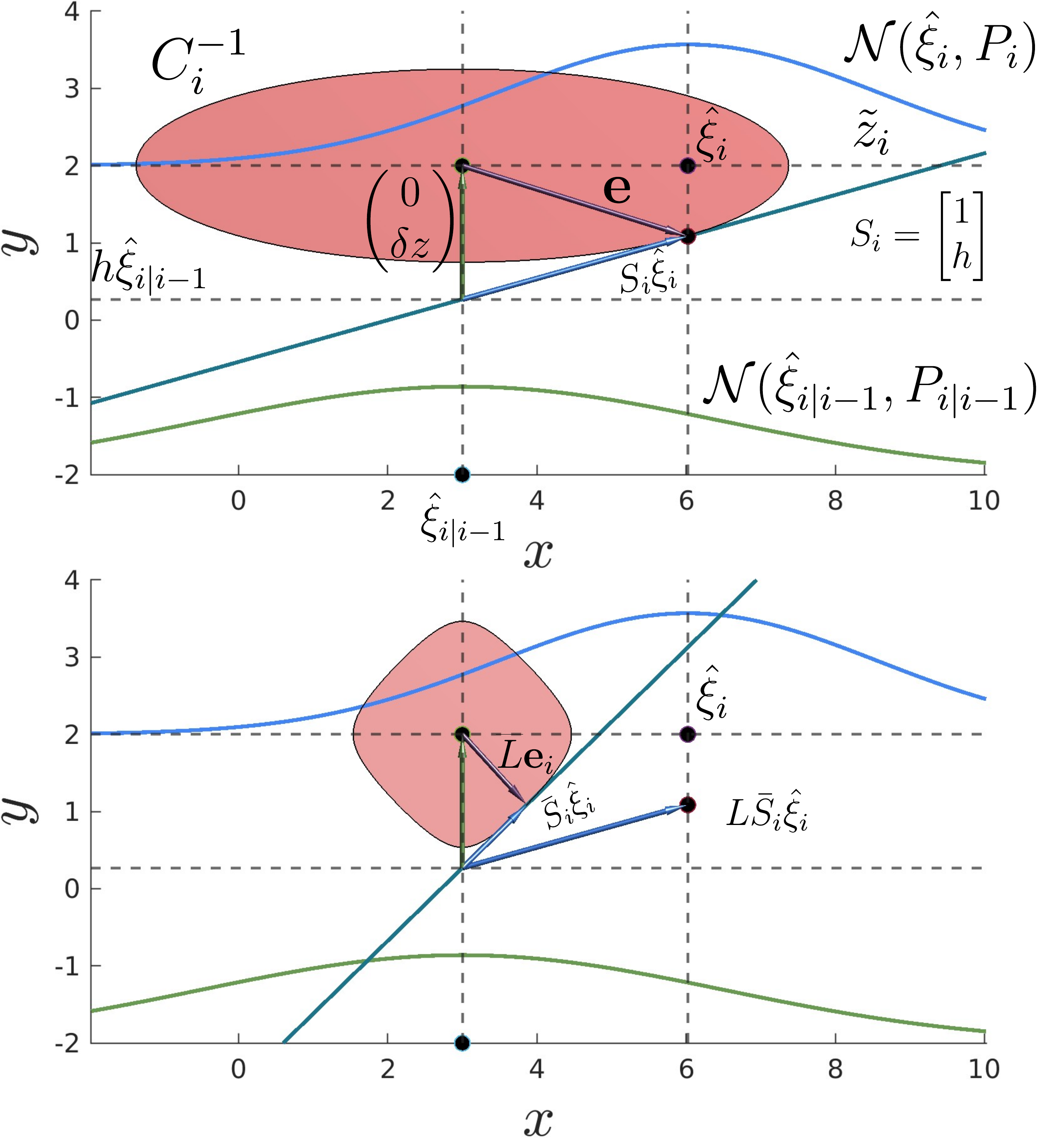}
    %\vspace{-1mm}
		\caption{The Kalman's measurement update for the one-dimensional case can be seen as a least square problem. The estimate can be seen as the intersection of $\mathbf{x}^{T} C^{-1} \mathbf{x}$ with $S_i$  (\ref{eq:robust_kalman}) (top).
		The same problem but changing $\rho$ to the Tukey cost function (c = 2.0). Note that a change of basis by $L$, $C_i = LL^{T}$, 
		should be performed \cite{zoubir2018} (bottom).}
    \label{fig:kalman_regression_robust} 
    %\vspace{-3mm}
\end{figure}

Although with optimal properties, the Kalman Filter is sensible to outliers. This happens because geometrically the Kalman Filter consists of the following linear regression problem (see Fig. \ref{fig:kalman_regression_robust}): 
\begin{equation}
\label{eq:robust_kalman}
\begin{aligned}
&\mathbf{b}_i :=\begin{bmatrix}
	\mathbf{0} \\
\delta \mathbf{z}_{W,i}
\end{bmatrix} = 
\underbrace{ \begin{pmatrix}
I \\
H_i
\end{pmatrix} }_{ S_i }{\boldsymbol{\xi}}_{r,i+1} + 
\underbrace{\begin{bmatrix}
\mathbf{n} \\
\mathbf{v}
\end{bmatrix}}_{\mathbf{e}_i} \implies \\&\hat{\boldsymbol{\xi}}_{r,i} = 
\min \rho(S_i {\boldsymbol{\xi}}_{r,i} - \mathbf{b}_i)
,\end{aligned}
\end{equation}
where $C_i = 
\begin{pmatrix}
P_{i|i-1} & \mathbf{0} \\
\mathbf{0} & \hat{N}_{i}
\end{pmatrix}$, $\mathbf{e}_i \sim \mathcal{N}(\mathbf{0}, C_i)$, $\rho(\mathbf{x}) = {\mathbf{x}^{T} C^{-1} \mathbf{x}}$. In this way,
since the gradient of $\rho$ is proportional to the size of $\mathbf{b}_i$, outlier measurements, e.g., slippage during locomotion, can strongly influence
the estimate $\hat{\boldsymbol{\xi}}_{r,i}$. To address this problem, we implemented the robust linear regression based on M-estimators (Algorithm 14 in \cite{zoubir2018}). This technique
substitutes $\rho(\cdot)$ in (\ref{eq:robust_kalman}) by the robust cost functions defined in (\ref{eq:huber}) and (\ref{eq:tukey}) which penalizes measurements greater
than a given threshold $c$. Intuitively, this technique substitutes the Mahalanobis norm with other convex norms as shown in Fig. \ref{fig:kalman_regression_robust} and the new optimization problem is solved 
by Iteratively Reweighted Least Squares (IRLS) method \cite{zoubir2018}. 
This same technique was used in \cite{das2021} to deal with wheel slippage in the context of planetary rovers. 

\section{RESULTS}
\label{sec:results}

The proposed methodology is tested with a public dataset from \cite{cerberus} called \textit{Street}, and with an indoor experiment using \textit{Unitree's} quadruped robot AlienGo. The methodology was implemented in C++ on Ubuntu 20.04 with ROS Noetic, and the tests were conducted with an AMD Ryzen 5 5600 CPU with 32GB of memory. We utilize the absolute trajectory error (ATE) and the relative pose error (RPE), defined in \cite{ate_rpe} as evaluation metrics.
\looseness=-1

\subsection{Street Dataset}
This dataset \cite{cerberus} includes sensor data collected from Unitree's A1 robot. The dataset comprises data from cameras, joints, IMU, and contact sensors, and it features an open-source Visual-Inertial-Leg Odometry (VILO) state estimation solution for legged robots. This solution accurately estimates the robot's position in real-time across various terrains using standard sensors such as stereo cameras, IMU, joint encoders, and contact sensors. For the A1 robot, we specifically used joint states, IMU measurements, and contact states. We compare our results with those obtained using the VILO state estimation.  Results are summarized in Table \ref{tab:cerberus}.

\begin{figure}[t!]
\centering
    \includegraphics[width=0.65\linewidth]{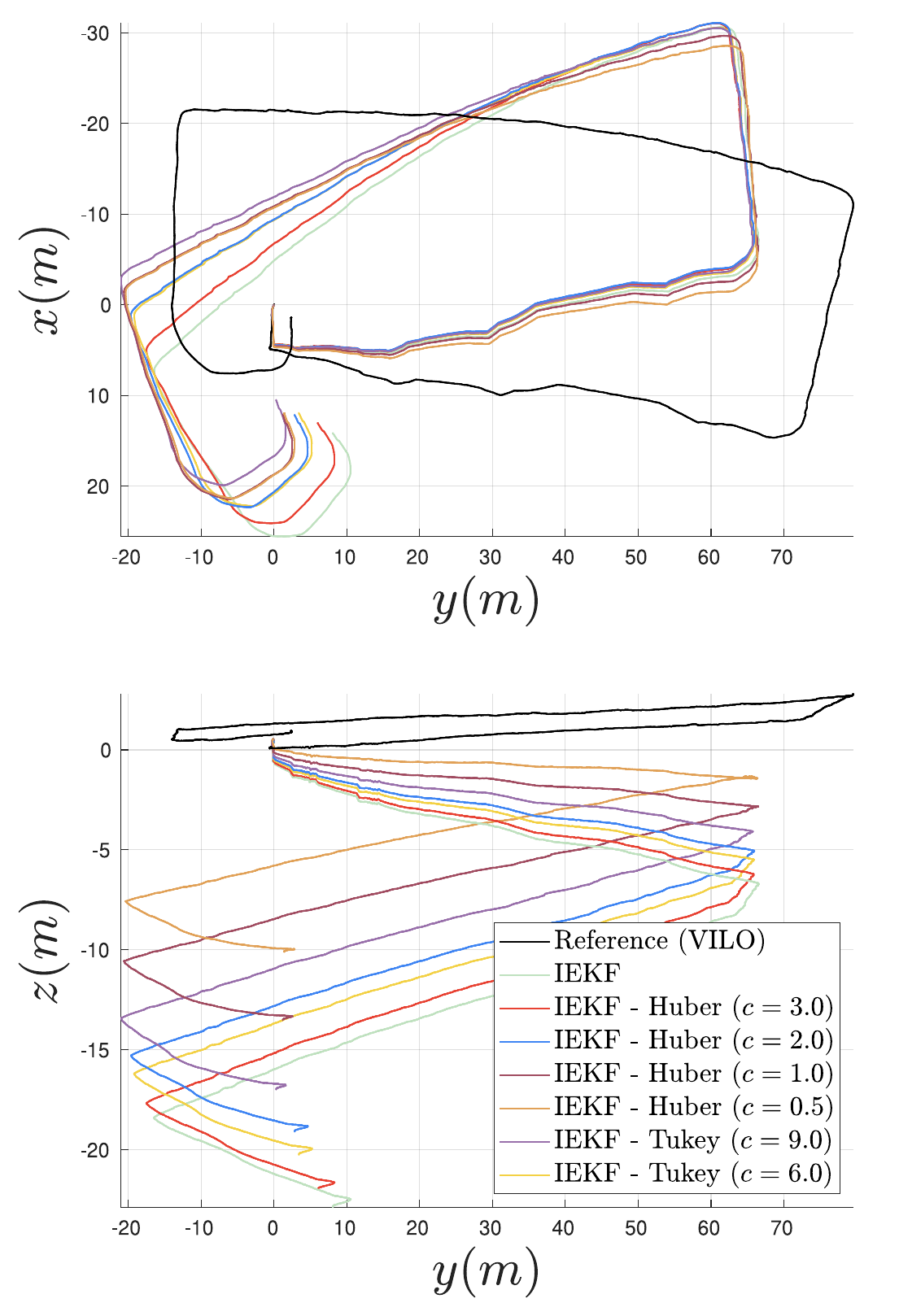}
    \caption{Comparison of the trajectory estimated by different methods of the dataset Street: X-Y plane (top); Z-Y plane (bottom).}
    \label{fig:cerberus_x-z} 
    \vspace{-5mm}
\end{figure}

\begin{table}[!b]
\begin{center}
\caption{ATE and RPE statistics over 1 m ($\sim$450 m trajectory) for the Street dataset}
\label{tab:cerberus}
    \begin{tabular}{c c c c}
        \toprule
        RMSE & ATE & RPE trans & RPE rot  \\
        \rowcolor{gray!15}
        \midrule
        IEKF & 9.237836 m & 0.657191 m & 6.959852$^\circ$ \\
        IEKF - Huber (c = 3.0) & 8.612143 m & \textbf{0.654514 m} & 6.960477$^\circ$ \\
        \rowcolor{gray!15}
        IEKF - Huber (c = 2.0) & 7.320663 m & 0.656467 m & 6.960724$^\circ$ \\
        IEKF - Huber (c = 1.0) & 5.947609 m & 0.661961 m & \textbf{6.956659$^\circ$} \\
        \rowcolor{gray!15}
        IEKF - Huber (c = 0.5) & \textbf{5.499099 m} & 0.663859 m & 6.960574$^\circ$ \\
        IEKF - Tukey (c = 9.0) & 7.574655 m & 0.655226 m & 6.958379$^\circ$ \\
        \rowcolor{gray!15}
        IEKF - Tukey (c = 6.0) & 6.352757 m & 0.656517 m & 6.958181$^\circ$ \\
        \bottomrule
    \end{tabular}
\end{center}
\end{table}

\subsection{AlienGo robot on uneven and slippery terrain}
\begin{figure}[!t]
\centering
    \includegraphics[width=1.0\linewidth]{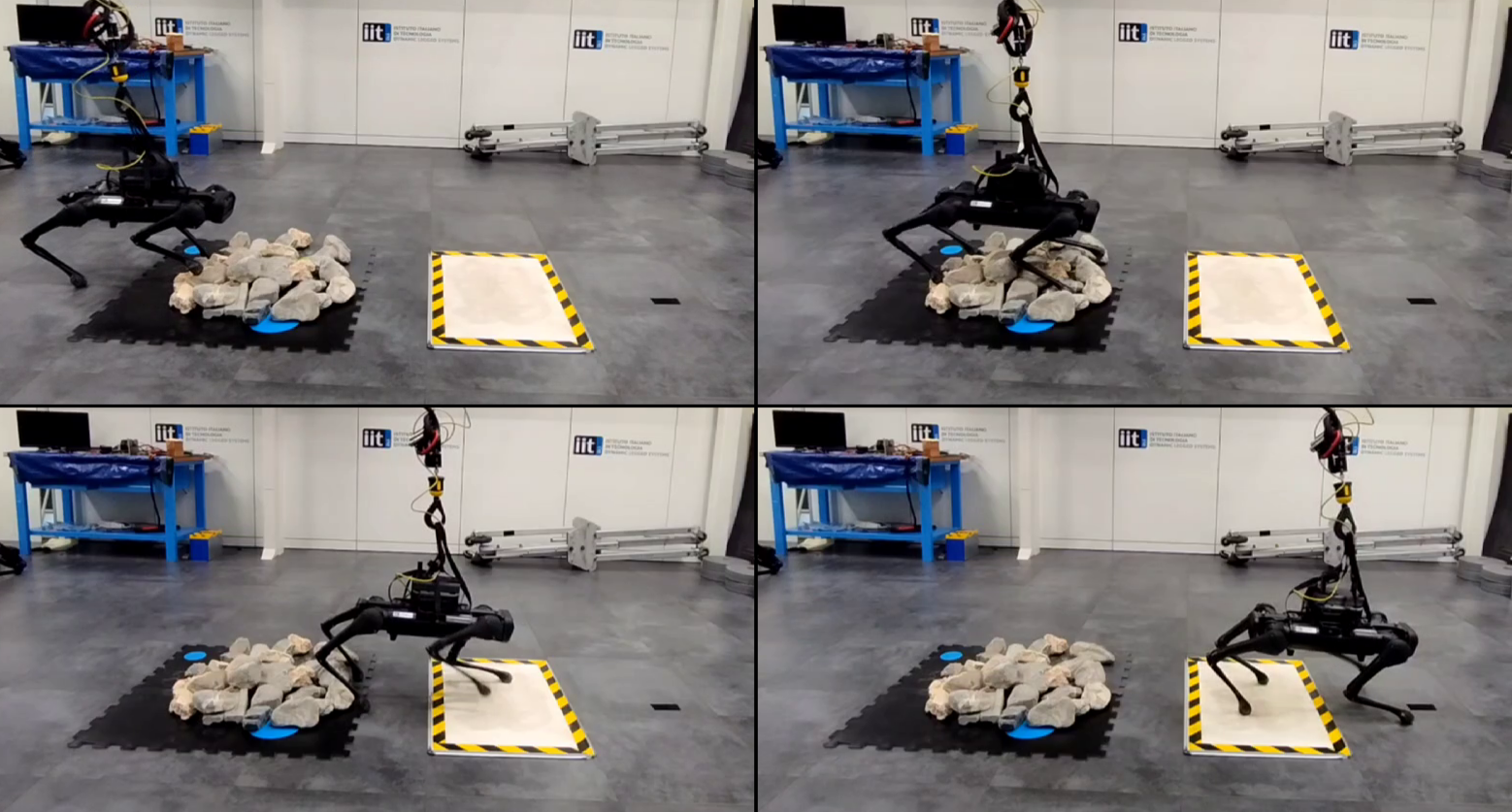}
    \caption{Aliengo walking on rocks and slippery terrain.}
    \label{fig:experiment_images} 
\end{figure}

In this experiment performed at IIT's DLS lab, the AlienGo robot used a crawl-gait to walk over uneven terrain composed of a pile of rocks followed by a white plastic sheet covered with liquid soap, as shown in Fig. \ref{fig:experiment_images}. The contact information is obtained by thresholding the ground reaction forces, similar to \cite{fink}. The ground truth was obtained using a \textit{Vicon} motion capture system. We compare our results with those obtained using the onboard RealSense T265's binocular visual-inertial SLAM, which is widely used in other works to aid state estimation \cite{teng2021icra}, \cite{vero}, \cite{bayer2019}. Results are summarized in Table \ref{tab:aliengo}.

\begin{figure}[t!]
\centering
    \includegraphics[width=0.7\linewidth]{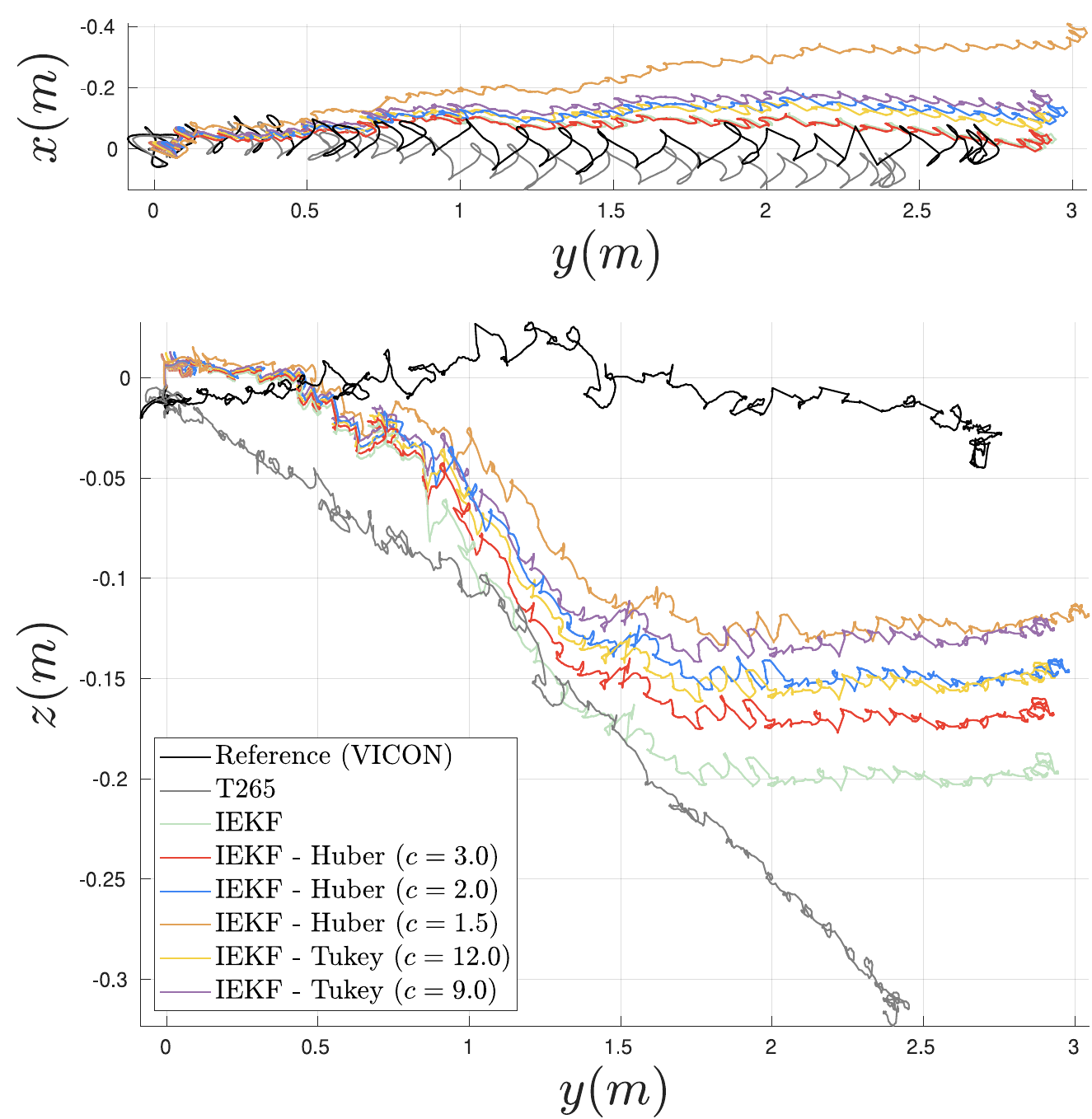}
    \caption{Comparison of the trajectory estimated by different methods of the experiment with Aliengo: X-Y plane (top); Z-Y plane (bottom).}
    \label{fig:experiment_zy} 
    \vspace{-4mm}
\end{figure}

\begin{table}[!b]
\begin{center}
\caption{ATE and RPE statistics over 1 m ($\sim$ 5 m trajectory) for the experiment with AlienGo}
\label{tab:aliengo}
    \begin{tabular}{c c c c}
        \toprule
        RMSE & ATE & RPE trans & RPE rot  \\
        \midrule
        \rowcolor{gray!15}
        T265 & 0.097843 m & \textbf{0.083525} m & \textbf{1.684767$^\circ$} \\
        IEKF & 0.076369 m & 0.124073 m & 7.484888$^\circ$ \\
        \rowcolor{gray!15}
        IEKF - Huber (c = 1.5) & 0.092125 m & 0.119336 m& 7.398602$^\circ$ \\
        IEKF - Huber (c = 2.0) & 0.077348 m & 0.121888 m & 7.469457$^\circ$ \\
        \rowcolor{gray!15}
        IEKF - Huber (c = 3.0) & 0.072696 m & 0.123331 m & 7.490033 \\
        IEKF - Tukey (c = 9.0) & \textbf{0.071265} m & 0.120438 m & 7.478264$^\circ$ \\
        \rowcolor{gray!15}
        IEKF - Tukey (c = 12.0) & 0.072393 m & 0.121869 m & 7.478770$^\circ$ \\
        \bottomrule
    \end{tabular}
\end{center}
\end{table}

\subsection{Discussion}
In both the dataset and experiment, we fixed the input covariance and varied the scale parameter $c$, using one of the two robust cost functions, Tukey or Huber.

The estimates for the Street dataset are shown in Fig. \ref{fig:cerberus_x-z}. The IEKF achieved the worst results in both X-Y and Z-Y planes. It is also shown that as the scale parameter decreases, it reduces the drift in both $X-Y$ and $Z-Y$ planes. This happens because as $c\to 0$ less influence is given by the outlier measurements. Moreover, it is shown that a Tukey cost function with $c =9.0$ has a similar result as a Huber with $c=2.0$.

We performed the same comparison as before for the experiment with AlienGo, i.e., we varied $c$ and the cost functions. It can be seen in Table \ref{tab:aliengo} that all robust runs reduced the IEKF's error, either in ATE or RPE, Tukey ($c=9$) being the best of them. Figure \ref{fig:experiment_zy} show the comparison between the IEKF, T265, and the robust IEKF with different cost functions. In this case, the increase in robustness caused a drift in the $X-Y$ plane and reduced the drift in the $z$ direction.
\looseness=-1

Analysing both cases, we can conclude that on a long trajectory without slippage where the drift is caused more by the action of the temporal propagation of the error, the increase in robustness reduces the drift in both planes. In a scenario with high slippage, on the other hand, the results indicate a trade-off between vertical drift and accuracy in the X-Y plane. Both cases clearly demonstrate the advantage of using the robust cost functions in the IEKF.

\section{CONCLUSIONS}
\label{sec:conclusion}

This paper presented a new method for proprioceptive state estimation of quadruped robots using an IEKF filter and scale-variant robust cost functions. We show in the methodology a new formulation of the IEKF for quadruped robots using recent theoretical formulations. We also demonstrated through dataset tests and experiments the advantage of using robust cost functions in the IEKF formulation in long and slippery trajectories.

In the analysis, we fixed the sensor covariance values. These values alter the outcome of the performance of the robust cost functions, but this was not explored in this paper. In future works, we aim to explore this, also focusing on finding optimal scale values.


%%%%%%%%%%%%%%%%%%%%%%%%%%%%%%%%%%%%%%%%%%%%%%%%%%%%%%%%%%%%%%%%%%%%%%%%%%%%%%%%

%%%%%%%%%%%%%%%%%%%%%%%%%%%%%%%%%%%%%%%%%%%%%%%%%%%%%%%%%%%%%%%%%%%%%%%%%%%%%%%%



%\section*{ACKNOWLEDGMENT}

%%%%%%%%%%%%%%%%%%%%%%%%%%%%%%%%%%%%%%%%%%%%%%%%%%%%%%%%%%%%%%%%%%%%%%%%%%%%%%%%%
\bibliographystyle{IEEEtran}
\bibliography{main}

\end{document}